\documentclass{article}
\usepackage{ijcai19}

\usepackage{booktabs}

\usepackage{color}

\usepackage{bbm}
\usepackage{float}
\usepackage{amsmath}
\usepackage{todonotes}
\usepackage[linesnumbered,ruled,vlined]{algorithm2e}

\DeclareMathOperator*{\argmin}{arg\,min}

\usepackage{graphicx}
\usepackage{subcaption}
\usepackage{bbm}
\usepackage{hyperref}
\usepackage{amsfonts}
\usepackage{multirow}
\usepackage{array}
\usepackage{changepage}
\usepackage{tabu}
\usepackage{supertabular}

\title{Exploring Computational User Models for Agent Policy Summarization}

\author{
Isaac Lage\footnote{Equal contribution}$^1$\and
Daphna Lifschitz$^*$$^2$\and
Finale Doshi-Velez$^1$\And
Ofra Amir$^2$\\
\affiliations
$^1$Harvard University\\
$^2$Technion - Israel Institute of Technology
\emails
isaaclage@g.harvard.edu,
daphna.l@campus.technion.ac.il,
finale@seas.harvard.edu,
oamir@technion.ac.il
}

\begin{document}

\maketitle

\begin{abstract}
AI agents are being developed to support high stakes decision-making processes from driving cars to prescribing drugs, making it increasingly important for human users to understand their behavior. Policy summarization methods aim to convey strengths and weaknesses of such agents by demonstrating their behavior in a subset of informative states. Some policy summarization methods extract a summary that optimizes the ability to reconstruct the agent's policy under the assumption that users will deploy inverse reinforcement learning. In this paper, we explore the use of different models for extracting summaries. We introduce an imitation learning-based approach to policy summarization; we demonstrate through computational simulations that a mismatch between the model used to extract a summary and the model used to reconstruct the policy results in worse reconstruction quality; and we demonstrate through a human-subject study that people use different models to reconstruct policies in different contexts, and that matching the summary extraction model to these can improve performance. Together, our results suggest that it is important to carefully consider user models in policy summarization.
\end{abstract}

\maketitle

\section{Introduction}
Autonomous and semi-autonomous agents are being developed and deployed to complete complex tasks such as driving or recommending clinical treatment. As these agents take a growing role in our daily lives, it is becoming increasingly important to provide ways for people to understand and anticipate their behavior. Recent works in the area of interpretability and explainable AI have thus developed methods for describing and explaining the decisions of agents to human users. A rich body of research focuses on explaining a \emph{specific} decision made by an agent to a human user. A recent complementary line of work focuses on \emph{summarizing} the \emph{global} behavior of the agent by demonstrating actions taken by the agent in different states~\cite{amir2018agent}. Such summaries have been shown to improve people's ability to assess agents' capabilities~\cite{amir2018highlights,huang2018establishing}, anticipate agents' actions~\cite{huang17communicate} and facilitate trust~\cite{huang2018establishing}. 

The problem of policy summarization, or extracting subsets of state-action pairs that globally characterize the agent's behavior, has been approached in two ways. The first approach applies heuristics related to the diversity or importance of states to determine what is shown in the summary~\cite{amir2018highlights,huang2018establishing}. The second approach assumes a computational model of how humans will generalize from the summaries provided, and uses that model to optimize the summary for reconstructing the agent's full policy \cite{huang17communicate}. Specifically, \cite{huang17communicate} assumed that people would deploy inverse reinforcement learning (IRL) to infer the agent's reward function from the summary; their summaries were created to perform well given this assumption of human computation. The cognitive science literature provides evidence that people sometimes do build models of others' behavior in this way~\cite{baker2009action,baker2011bayesian}, making IRL a plausible model. However, there also exists evidence that human planners sometimes rely on a model-free system that computes actions based on previous experience in a situation~\cite{daw2005arbitration}. People may do something similar when inferring the behavior of others, making imitation learning (IL) another plausible model of human computation. 

In this paper, we explore the effects of using different models for summary extraction on the ability to reconstruct a policy. We make the following contributions: (1) we develop an IL-based summary extraction method; (2) Through computational simulations in a variety of domains, we demonstrate that the model used during summarization needs to match the model used to reconstruct the policy to produce high quality reconstructions; and (3) we demonstrate through human-subject studies that people may deploy different models in different contexts and that in some cases matching summary extraction to reconstruction model results in improved policy reconstruction. Taken together, these results suggest the importance of carefully considering which computational models of users we employ during policy summarization.

\section{Related Work}

\textbf{Summarizing Agent Policies.} Policy summarization is the problem of extracting a collection of state-action pairs that globally characterize an agent's policy with the goal of helping a  user  understand it~\cite{amir2018agent}. One approach relies on heuristics based on agent Q-values and state similarity, to extract diverse, important states~\cite{amir2018highlights,huang2018establishing}. A second approach formalizes the problem through machine teaching, where the goal is to extract state-action pairs useful for recovering the agent's reward function with IRL~\cite{huang17communicate}. Our work extends the latter approach by additionally considering IL.

\textbf{Cognitive Science of Inferring Agent Behavior.}
\cite{dragan2014familiarization} show that people can learn to anticipate agent actions if they see enough examples, but some agent behaviors can never be fully anticipated. Baker et al.~\shortcite{baker2009action,baker2011bayesian} suggest that people use ``theory of mind'' to infer others' beliefs and desires based on observations of their actions, modeling people's inference as Bayesian inverse planning. \cite{daw2005arbitration} describes how human planning uses both a model-based system that computes rewards and transitions, and a model-free system that computes actions based on previous experience. While in their setting people learn by directly interacting with the world, a similar set of approaches may also be used when inferring actions of others. Work on agents modeling agents \cite{albrecht2018modeling} describes strategies including case-based reasoning, of which IL is an example, and utility reconstruction, of which IRL is an example. Finally, \cite{medin19778context} discusses the exemplar theory of classification in psychology that says that humans categorize new objects based on similarity to objects in their memory, motivating our choice of IL model.

\textbf{Explaining Agent Decisions.}
More broadly, our work relates to the area of explainable AI~\cite{aha2017ijcai}. Many approaches focus on explaining a specific decision \cite{khan2009minimal,lomas2012explaining,dodson2011natural,broekens2010explain} or showing which features the agent attends to~\cite{greydanus2017visualizing}. \cite{miller2018explanation} reviews findings from the social sciences regarding useful explanations, and defines an explanation as causal, providing an answer to a why-question. In contrast to these approaches, we consider a complimentary non-causal \emph{description} of the agent's global behavior. Closer to our work, \cite{vanderwaa2018contrastive} explains policies in terms of differences in expected outcomes from a user specified, contrast policy, and \cite{ramakrishnan2018blindspot} formalizes the problem of detecting ``blind spots'', situations in which an agent acts incorrectly because it cannot differentiate between important real world states. Our work aims to more generally provide a summary that highlights the agent's strengths as well as weaknesses without requiring the user to be able to specify a specific contrast.

\section{Methods}
\label{sec:methods}

Following \cite{huang17communicate}, we formalize the problem of extracting a summary of an agent's behavior as an instance of machine teaching, which aims to find a set of training examples that induces a known target model to learn a pre-specified source model~\cite{zhu2015machine}. In our setting, the agent's policy is the source model that we want to induce, and the target models that we consider are hypotheses about how humans generalize from examples of an agent's behavior---IRL and IL.
 
Formally, our problem is to find the set of examples $T$ of size $k$, where $T = \langle \langle s_{1},a_{1} \rangle,...,\langle s_{k},a_{k} \rangle \rangle$ is the set of state-action pairs that maximizes the similarity $\rho$ of the policy induced by $T$ under a specific target computational model $M$. The objective can be written as:
\begin{equation}
 \max_{T \in \mathcal{D}} \rho( \hat{\pi}(T,M), \pi^*)\\
 \texttt{s.t.} |T| = k
\end{equation}
where $\mathcal{D}$ is all state-action pairs demonstrating the agent's policy, $\pi^{*}$, $\hat{\pi}(T,M)$ is the approximate policy attained by applying computational model $M$ to the summary $T$, and $\rho$ is a measure of similarity between the agent's true policy $\pi^{*}$ and the reconstructed policy $\hat{\pi}$.

\subsection{IRL-Based Summary Extraction}
Given a collection of trajectories, IRL extracts a reward function such that the optimal policy with respect to those rewards matches the demonstrated behavior \cite{ng2000algorithms}. This captures the notion that people may first infer the agent's reward function, then use it to replicate the agent's planning process.

\paragraph{Model of Human Extrapolation: Maximum Entropy IRL.}
Max-Ent~\cite{ziebart2008maximum} is a model-based approach to IRL that formulates the problem of learning a policy from observed trajectories as optimizing a linear function, mapping the features of each state to a reward value. Its goal is to match the \emph{feature expectations} of the learned policy to those of the observed trajectories. These expectations are defined as 
\begin{equation}\label{eq_safcount}
 \mu_\pi^{(s,a)} = \mathbb{E}[
 \sum_{t=0}^{\infty} \gamma^t\phi(s_t)|\pi, s_0=s, a_0=a]
\end{equation} 
Where $\mu_\pi^{(s,a)}$ are the feature expectations resulting from starting at state $s$, taking action $a$ and following the policy $\pi$ thereafter. $\phi(s_t)$ is the feature vector of state $s_t$ and $\gamma$ is a discount factor. There may be many reward functions resulting in feature expectations that match the observed trajectories; Max-Ent chooses one based on the maximum entropy principle. We note that Max-Ent results in the same computations as the probabilistic reward-based model used for IRL-based summary extraction in Huang et al.~\shortcite{huang17communicate}, even though the former assumes possible noise in expert demonstration and the latter assumes noise in the reconstruction.

\paragraph{Summary Extraction Method: Machine Teaching.}
We produce a summary of a policy by extracting trajectories that maximize the quality of its reconstruction of the policy using Max-Ent. We do this using the SCOT machine teaching algorithm~\cite{brown2018machineteachingirl} that selects a minimal set of demonstrations which allows the learner to obtain a reward function \textit{behaviorally equivalent} to the optimal policy $\pi^*$. The behavioral equivalence class (BEC) of $\pi^*$ is defined as the set of reward functions under which the policy is optimal. The BEC of $\pi^*$ can be expressed by the intersection of halfspaces given by the following constraints:
\begin{equation}\label{eq_BEC}
 w^T(\mu_{\pi^*}^{(s,a)} - \mu_{\pi^*}^{(s,a')}) \geq 0, 
 \forall \langle s , a \rangle \in \mathcal{D}, \forall a' \in A
\end{equation}
where $A$ is the set of actions available to the agent, $w\in\mathbb{R}^k$ are the reward weights, and $\mu_{\pi^*}^{(s,a)}$ are the expected feature counts as described above. The BEC of a demonstration is defined by the intersection of halfspaces for the demonstrated states and actions. SCOT greedily finds the smallest set of trajectories with halfspace constraints covering the constraints defined by $\pi^*$. Here, $\rho$ is infinite if the reconstructed reward function belongs to the BEC of $\pi^*$ and 0 otherwise.

We modified the algorithm to extract a fixed budget $k/l$ of trajectories, where $k$ is the number of states in the summary and $l$ is the trajectory length. We do so by terminating after the budget is reached, or by randomly adding additional trajectories when the budget is larger than the number of trajectories required to cover the set of non-redundant constraints.

\subsection{IL-Based Summary Extraction}
Given a set of states and actions, IL learns a function $\hat{\pi}: s\,\to\,a$ mapping directly from states to actions. This captures the notion that people may predict the agent's action based on actions in similar states, with no concept of reward or goal. 

\paragraph{Model of Human Extrapolation: Gaussian Random Field.}

The GRF model in \cite{zhu2003} represents data points---in our case, states---as vertices in a graph connected by edges weighted by their similarity. It makes predictions by propagating labels---in our case, actions---through the graph. In the binary setting, the action probabilities can be written as follows:
\begin{equation}\label{eq_energy}
p(\mathcal{D}) = \frac{1}{Z_\beta}\exp(-\beta (\frac{1}{2} \sum_{\substack{\langle s , a \rangle \in \mathcal{D}, \\ \langle s' , a' \rangle \in \mathcal{D}}} v(\phi(s), \phi(s')) (a - a')^2))
\end{equation}
 where $v$ is a kernel, $\beta$ is a tunable inverse temperature parameter (we set $\beta=1$) and $Z_\beta$ is a normalizing constant. We extend this to the multiclass setting with one-vs-rest classification as suggested in \cite{zhu2003semisupervised}. 

Predictions are made as follows: 
\begin{equation}\label{eq_GRFmean}
\hat{\pi}_U = - L_{UU}^{-1} L_{UT} \hat{\pi}_T
\end{equation}
where $u = \mathcal{D} \setminus T$, $\hat{\pi}_T = \pi^*_T$, and $L = diag(\sum_{\langle s' , a' \rangle \in \mathcal{D}} v(\phi(s), \phi(s'))) - V$ is the combinatorial Laplacian matrix where $V$ represents the matrix of $v(\phi(s), \phi(s'))$ for all pairs of states. Predictions are binarized by thresholding at 0.5.

\paragraph{Summary Extraction Method: Active Learning.}
As with the IRL approach, given this model of human extrapolation, we need to define a procedure for producing a policy summary by extracting trajectories that maximize the quality of its reconstruction of the policy. We do this with the active learning algorithm in \cite{zhu2003} modified to account for the fact that we know ground truth values of $a$ for ${\langle s , a \rangle \in U}$. The algorithm implements the expected error reduction strategy, greedily choosing at each step to include the state-action pair, $\langle s , a \rangle^*$, that minimizes the 0/1 loss on all unseen states (which is $\rho$ in this case):
\begin{equation}\label{eq_ALloss}
\langle s , a \rangle^* = \argmin_{\langle s , a \rangle \in U} \sum_{\langle s' , a' \rangle \in U \setminus \langle s , a \rangle} \mathbbm{1}_{a' = \hat{\pi}^+(s')}
\end{equation}
where $\hat{\pi}^+$ is the model that has been retrained with $\langle s , a \rangle$ added into the training set. The GRF allows for efficient re-training of $\hat{\pi}^+$. 

\section{Computational Experiments}
We conducted computational experiments to address the question of whether the model used to extract a summary needs to match the hypothesized model used by humans to reconstruct the policy in order  to produce high-quality reconstructions.
We extract summaries using both the IL and IRL models of human extrapolation described in Section~\ref{sec:methods} 
and measure reconstruction quality for each summary with both models.

\subsection{Empirical Methodology}

\paragraph{Domains.}
We used three diverse domains.

\textit{Random Gridworld}: We use a 9x9 random grid world similar to the one described in \cite{brown2018machineteachingirl} as an example of a static, navigational environment. 
We use a 5-D one-hot feature vector and draw the rewards for each indicator without replacement from [100, 10, 0, -10, -100]. The policy is determined with value iteration using a discount factor of $0.95$. 

\textit{PAC-MAN}: We use a 6x7 PAC-MAN grid with a single food pellet in the middle, a wall surrounding it on 3 sides and a single ghost that moves towards PAC-MAN deterministically as an example of a dynamic, navigational environment.\footnote{\url{http://ai.berkeley.edu/project_overview.html}} The policy takes PAC-MAN to the nearest food that does not result in a ghost collision. 

\textit{HIV Simulator}: We use the HIV simulator described in \cite{adams2005hiv} which includes 6 biomarker features, and 4 actions corresponding to activating or not activating 2 drugs. This domain serves as an example of a non-navigational, signal-based environment. The policy is determined with fitted Q iteration as in \cite{ernst2006clinical} with a 0.05 initial state perturbation.

We describe state representation design choices including discretization of the the HIV domain and augmentation of the feature set with neighboring states for IL for gridworld and PAC-MAN 
in Appendix~\ref{sec:computational_details}.

\paragraph{Reconstruction Quality Measures.}
We use two metrics for reconstruction quality: the accuracy of predictions on states not included in the summary, and the absolute difference between the value of the original policy and that of the reconstructed policy. We include both measures because the IL summarization method optimizes the first and the IRL summary method indirectly optimizes for the second as its similarity $\rho$. We computed the accuracy over all unique, unseen states in the random gridworld and PAC-MAN domains, and over the unseen states from a batch of 5 episodes of 200 steps from the HIV simulator\footnote{5 episodes were sufficient to capture variation.}. We computed the value for the random gridworld and PAC-MAN domains using a single simulation of length 10 (both domains are deterministic) from each state with a uniform distribution over start states, and for the HIV simulator over 5 episodes of 200 steps starting from the initial state. 

\paragraph{Method Details.} To determine the summary size $k$ and the hyperparameters in the extraction model for each domain, we computed 75 random restarts of each hyperparameter settings' reconstruction quality from a summary extracted with its matched model. We chose the smallest summary size such that increasing it does not result in changes in the best performing methods for either IL or IRL in the HIV simulator and PAC-MAN domains. In the random gridworld domain, increasing the summary size always improved IL performance, so we choose a summary size such that the best performing IRL methods did not change (HIV: 24; Gridworld: 24; PAC-MAN: 12). We report results only for the best performing methods for IL and IRL at the chosen summary size. We searched over summary sizes [12, 24, 36, 48, 60]; IL hyperparameters: kernel [RBF, polynomial], length scale [0.1, 1.] and degree [2, 3] (for polynomial kernel only); and IRL hyperparameters: trajectory lengths [1, 2, 3, 4]. See Appendix Figure~\ref{fig:hyperparameters}. Max-Ent requires specifying additional hyperparameters that we held fixed (see Appendix~\ref{sec:computational_details}).

\subsection{Results}

\begin{figure*}
\centering
\includegraphics[width=0.75\linewidth]{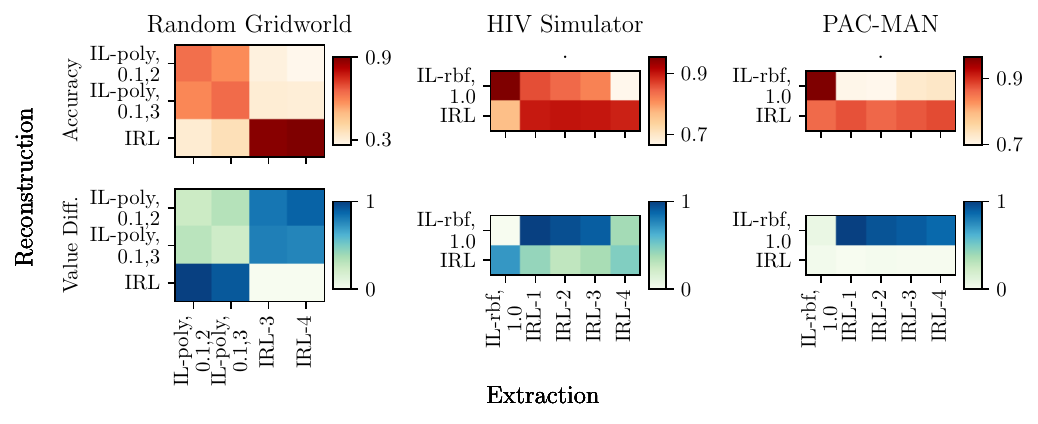}
\caption{Accuracy (higher is better) and 0-1 scaled value differences (lower is better) averaged over 75 random restarts of every reconstruction model (rows in heatmaps) used for summary extraction with summaries extracted with each model (columns in heatmaps). IRL hyperparameter corresponds to trajectory length; IL hyperparameters indicate kernel, length scale, degree in that order. 
The comparatively high accuracy and low value differences for the reconstructions where the summarization model matches the reconstruction model, indicate that matching the summarization model to the reconstruction model is important for producing a high quality reconstruction.}
\label{fig:computationalResults}
\vspace{-0.2cm}
\end{figure*}

Figure~\ref{fig:computationalResults} shows the accuracies and the 0-1 scaled value differences between the original policy and the reconstructed policy (raw values are not easily interpretable) for the different reconstruction models (rows) and different summary extraction models (columns).

\paragraph{Different reconstruction models result in higher absolute quality reconstructions in different domains when summaries are optimized for them.} The first question we considered was whether one approach---IL or IRL---always produced better summaries. In the HIV simulator and PAC-MAN domains, the IL reconstructions with IL summaries have higher accuracy than the IRL reconstructions with IRL summaries. In the HIV simulator domain, this is reflected in the value difference results, while in the PAC-MAN domain, the methods perform similarly with respect to value difference. In the random gridworld domain, the IRL reconstruction with the corresponding IRL summary has a higher accuracy and lower value difference than IL reconstructions with the corresponding IL summaries. These results indicate that different reconstruction models are more effective in different domains (given a summary optimized for that model). This effect likely has to do with how well each computational model can capture the policy. In the random gridworld, for example, the IRL model can perfectly model the policy but the IL model lacks important spatial information. 

\paragraph{Matching the extraction model to the reconstruction model is the most effective strategy for producing high-quality reconstructions.} Highest quality reconstructions occur when the same model is used for summary extraction and policy reconstruction. This is true even when a particular reconstruction model was generally more accurate. There are two exceptions to this: the IRL reconstruction with IL summary in PAC-MAN, that performs comparably in accuracy and value difference to the IRL reconstruction with IRL summary, and the value difference for the IL reconstruction with the IRL length 4 summary in HIV which is low even though action prediction accuracy is low. In both these cases, the reconstruction quality is no worse than with the summary based on the matched reconstruction model, and in the other cases it is much better. These results indicate that both the IL and IRL summarization methods are effective when the reconstruction model is known, and that using the correct reconstruction model during summarization can be very important for policy reconstruction quality.

\section{Human-Subject Study}

We conducted a human-subject study to test if the findings from our computational simulations generalize to humans, and to examine which reconstruction models people will naturally deploy.

\subsection{Empirical Methodology}

\paragraph{Task.} Participants inspected a summary of an agent's policy, based on which they were asked to predict the agent's actions in a subset of states not included in the summary. This parallels the accuracy measure of reconstruction quality used in the computational experiments, and tests how well people can generalize an agent's behavior from a summary of its policy.

\textit{Domains.} We used the HIV simulator and random gridworld domains because we expected people to have fewer priors about these than PAC-MAN (we disguised the disease, medication and biomarker names for HIV), and because understanding the transition function is more complex in the HIV domain, which we hypothesized would reduce people's ability to use IRL.

\textit{Summaries.} For each domain, we chose one summary each for IL and IRL where the pattern of higher accuracy for matched-summary reconstructions from the computational experiments was maintained. For gridworld, we showed length 24 summaries (IL: polynomial kernel, length scale=0.1, degree=2; IRL: trajectory length=4). For HIV, we showed length 12 summaries (IL: RBF kernel, length scale=1.0; trajectory length=3).\footnote{The HIV summaries are shorter than in the computational experiments so that they will fit on a single page, but this does not affect the trends from the computational results.}

\textit{Prediction States.} We selected test states to satisfy 2 criteria: 1) A similar computational accuracy grid to the full set of unseen states. 2) A uniform distribution over actions taken by the policy. For the random gridworld we selected states that do not have multiple optimal actions in the policy to avoid ambiguity. We describe additional subtleties in Appendix~\ref{sec:user_details}.
 
\paragraph{Design.} Our experiment used a 2X2 between subject design with domain [HIV or gridworld] and summary type [IRL or IL] as the main factors.

\paragraph{Participants.} 147 participants (Gridworld: IL=36, IRL=39; HIV: IL=37, IRL=35) were recruited through Amazon Mechanical Turk (65 female, Mean age = 36.38). 
Participants received a base payment of \$1.5, and a bonus of up to \$1 based on their performance.\footnote{The study was approved by our IRB.}

\begin{figure}[h]
\centering
\includegraphics[width=1\columnwidth]{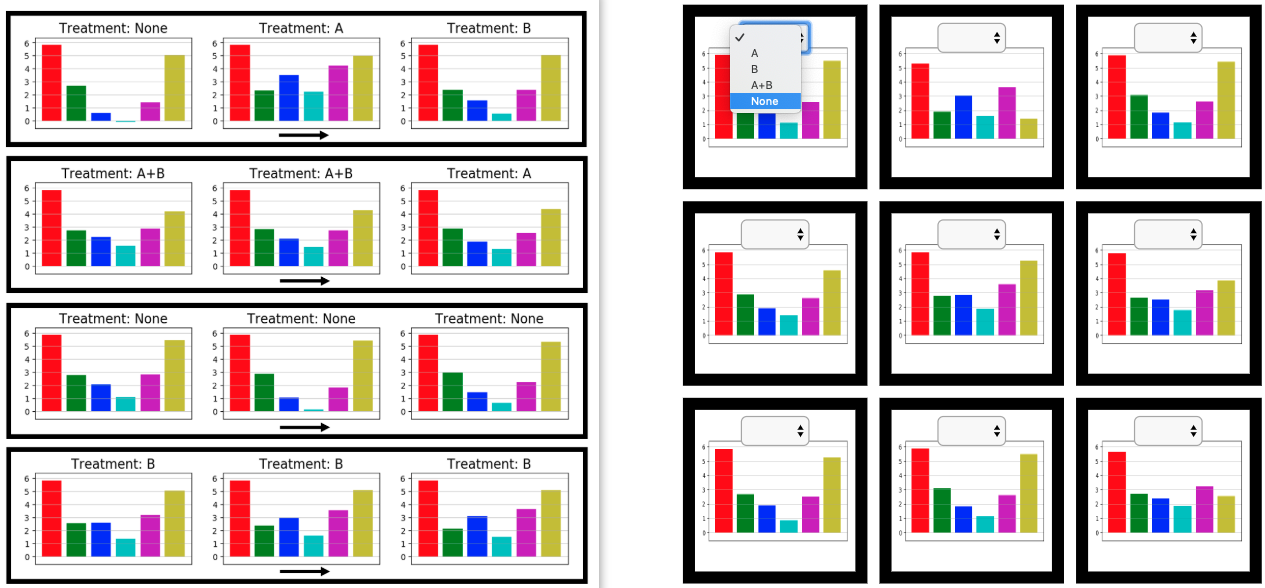}
\caption{The user study interface for an IRL summary in the HIV simulator domain. The left side shows the summary and the right side shows the prediction states.}
\label{fig:interface}
\vspace{-0.3cm}
\end{figure}

\paragraph{Interface.} Participants completed all predictions on the same page by choosing actions from dropdown menus with the summary displayed on the left. For the HIV simulator, this consisted of either a grid of independent states each outlined in black for IL, or a grid of trajectories outlined in black for IRL. States were visualized as bar graphs with the action written above (see Figure~\ref{fig:interface}). For the random gridworld, this consisted of the full grid\footnote{We argue this difference is a feature of the domains since it is not obvious how to display the full domain for HIV.}, with actions represented as arrows, and trajectories represented by different colors. See Appendix~\ref{sec:user_interface} for screenshots of all task interfaces.

\paragraph{Procedure.} 
Participants were given instructions explaining the domain and the task, following which they had to pass a quiz ensuring they understood the instructions. Upon passing the quiz, they were shown the summary and answered a practice round predicting 3 actions. Then, they completed the main task and predicted 9 additional actions. After making all predictions, participants were asked to provide a brief text description of how they made their predictions.

\subsection{Results}
\paragraph{In the HIV domain, most people employed IL-based reconstruction models, and performance was better with IL summaries.} Based on the qualitative responses to how people made their predictions, 1 study author coded participants as doing IL, IRL or not obviously doing either and a second author verified the coding (see Appendix~\ref{sec:user_qualitative}). In this domain, 78\% of participants reported using IL-based methods for reconstruction (e.g. ``I chose based on the similarity of the blood tests levels from the scenarios on the left''), while only one participant reported using an IRL-based method (``Treatment A is used to decrease middle ones(blue, light blue and purple)...''). Participants who were shown the IL summary performed significantly better compared to those shown the IRL summary with mean accuracy of 0.45 for participants shown an IL summary and 0.33 for those shown an IRL summary (Mann–Whitney U=330.0, n1=37 , n2=35, $P<0.001$ two-tailed). This demonstrates that (1) there are cases where people use IL-based reconstruction models; and (2) matching the computational model used during summarization to people's reconstruction model can improve reconstruction quality, paralleling our simulation results.

\paragraph{In the gridworld domain, people varied in the reconstruction models they employed showing a tendency towards IRL, and there was no difference in performance for different summary types.} In the random gridworld domain, 15\% of participants described IL-based reconstruction (e.g. ``I tried comparing the colors and deciding which was more frequent for a color.''), while 27\% provided descriptions suggesting IRL-based reconstruction (e.g. ``I decided that the computer seems to be always working towards a blue square. I chose the simplest path to get to a blue square''). The remainder of descriptions were too vague to imply a specific method. In contrast to the computational experiments, in this domain there was no significant difference in participants' performance based on summary type with mean accuracy of 0.38 for participants shown an IL summary and 0.37 for those shown an IRL summary (Mann–Whitney U=636.5, n1=36 , n2=37, P=0.24 two-tailed). Participants who reported using IRL reconstruction did perform significantly better than those who did not with mean accuracy of 0.27 for participants who mentioned using IL reconstruction and 0.66 for those who mentioned using an IRL reconstruction (Mann–Whitney U=188.0, n1=11, n2=20, $P<0.001$ two-tailed) as predicted by the computational results, but there was again no difference in performance between summaries among those who used IRL reconstruction. 

\paragraph{There are important differences between computational and human reconstructions including different feature spaces and lower accuracies.} Overall, participants' reconstruction accuracy was lower than predicted by matching the extraction and reconstruction models in the computational simulations, though much higher than random guessing (HIV: 0.33-0.45 compared to 0.67-0.78 for matched and 0.25 for random; Gridworld: 0.37-0.38 compared to to 0.78-1 for matched and 0.2 for random). In the random gridworld domain, reconstruction accuracies were also higher than predicted by mis-matched extraction and reconstruction models (0.37-0.38 compared to 0.11-0.22 for mis-matched). In the qualitative responses for the gridworld condition, some participants described the agent's behavior as based on absolute position (e.g. ``It seems like the agent is trying to get to the center of the screen''), despite being told the agent navigates only based on tile color. In HIV, some people relied on absolute values of the features, and others tried matching the ``shape'' of the bar graphs (e.g. ``I compared the levels of each of the colors and the shapes of the graphs.'') Perhaps due to these disconnects in feature spaces and possible cognitive limitations, people had worse reconstruction accuracy than the computational models with matched summaries; however, these tendencies may have also enabled the better accuracies we observed in the random gridworld domain with mismatched summaries. 

\section{Discussion \& Future Work} 

In this paper, we explored how the computational models of users employed during policy summarization affect people's ability to reconstruct an agent's policy. Computational simulations in 3 diverse domains showed the importance of matching the summarization model to the reconstruction model. Human-subject studies showed that people use different models when reconstructing policies, sometimes deploying IL and sometimes IRL, and that the model used during summarization sometimes affected the quality of their reconstructions. Together, these findings demonstrate the importance of personalizing user models for summarization to domain context.

Our findings suggest several avenues for future work. First, future studies can explore in which circumstances people use different reconstruction models. We hypothesize that familiarity with the domain might be one aspect, as we observed that in the less familiar domain of HIV treatment, people tended toward IL. Better understanding of when people use which approach can help ensure that matching models will be used for extraction and reconstruction.

Second, there are additional nuances to people's extrapolation beyond the choice of IL or IRL -- e.g., which feature representation they use, and how exactly they perform inference. Given the sensitivity of the ability to reconstruct policies to summary extraction models, and the finding that different people use different models even within the same domain, we argue that human-in-the-loop approaches to summary extraction are a promising approach to ensure that summaries match the user's reasoning about a particular domain.

Finally, approaches to policy summarization based on models of user extrapolation can be integrated with those based on ``important'' states or identifying blind-spots. All of these can then be combined with approaches for \emph{explaining} specific agent decisions, to produce summaries that simultaneously demonstrate global agent behavior and explain specific actions. 

\subsubsection{Acknowledgements} 
The authors acknowledge a Google Faculty Research Award and the J.P. Morgan AI Faculty Research Award. IL is supported by NIH 5T32LM012411-02.

\small
\bibliographystyle{named}
\bibliography{bibliography}

\appendix

\section{Computational Experiment Details}
\label{sec:computational_details}

\paragraph{State Representations.} 
In the PAC-MAN and random gridworld domains, we used different feature sets for IL and IRL so that IL would have access to some short-term temporal information. In PAC-MAN, we use distance to nearest food, an indicator for food consumption, and an indicator for being eaten by a ghost as features for the IRL model, and the direction of nearest food, an indicator for collisions with ghosts or walls in each direction for the IL model. In the random gridworld, we used the 5-dimensional vector as features for the IRL model, and we additionally used the concatenated vector for each neighboring tile for the IL model. In the HIV simulator domain, we used the 6 biomarkers as features for both IL and IRL. 

The continuous HIV simulator domain, required discretization to build the explicit transition function required by the IRL model. To do this, we ran K-Means clustering with 100 clusters on each state in the extracted trajectories, and used the cluster centers as the state representations.\footnote{100 clusters resulted in accuracy above $0.95$ when using the most common action in each cluster for prediction.} We use this representation for the IRL reconstruction and to compute the value, but otherwise we use the original state representations.

\paragraph{Fixed Max-Ent Hyperparameters}

We held the following choices for the IRL model fixed. We set the discount factor to $\gamma = 0.95$ for Random Gridworld, $\gamma = 0.98$ for HIV to match the discount factor used to derive the policy.  For PAC-MAN, the policy was derived without a discount factor so we set it to $\gamma = 0.95$. In the Random Gridworld and PAC-MAN domains, we set the rollout horizon to 10, while in the longer time-horizon HIV domain we set the rollout horizon to 25. For Max-Ent, we set the learning rate to 1 for the random gridworld, 0.1 for PAC-MAN and 0.01 for the HIV simulator and ran 100 iterations, stopping if the rewards changed by less than 1e-5 between two consecutive iterations.

\section{Hyperparameters}
\label{sec:hyperparameters}

\begin{figure}[h]
\centering
\includegraphics[width=1\columnwidth]{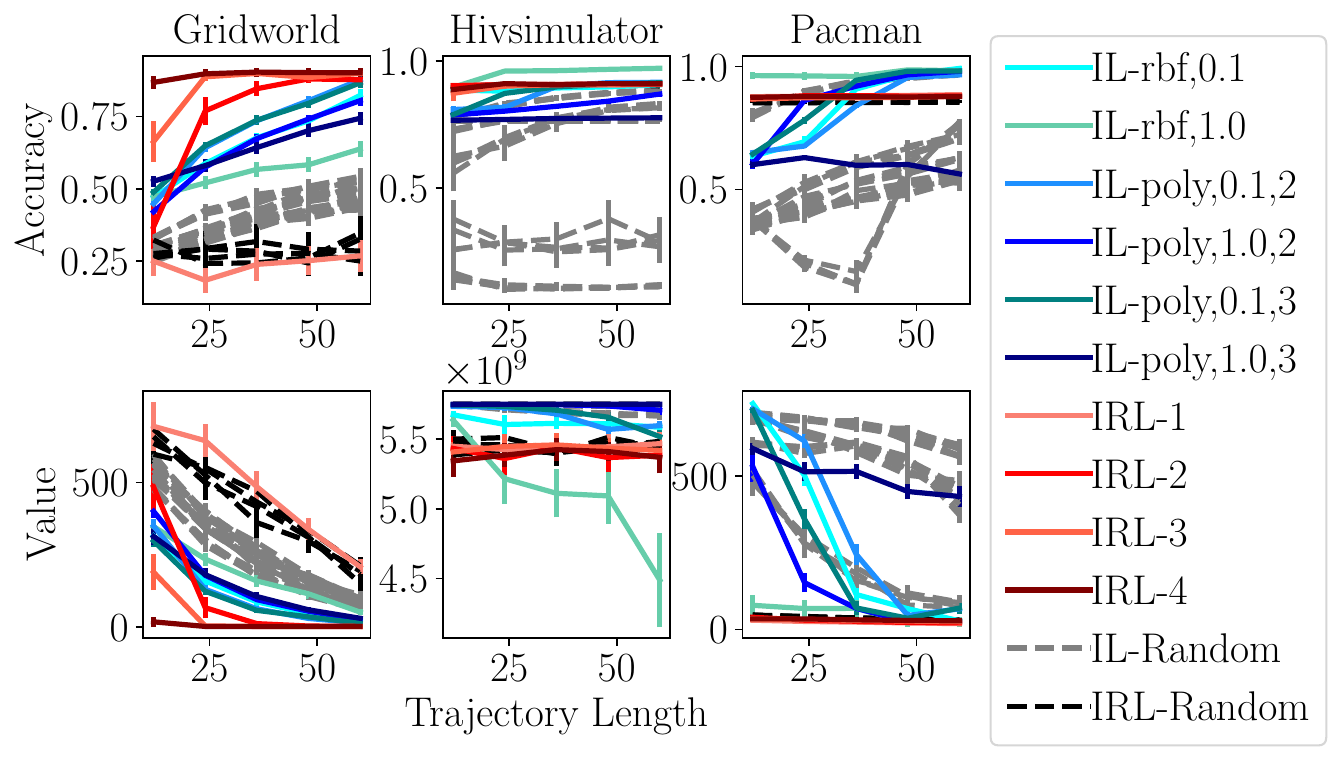}
\caption{Reconstruction accuracy across a variety of reconstruction methods and hyper-parameter settings (assuming the same extraction method), by summary size, averaged over 75 random restarts. Error bars signify 95\% confidence intervals. We chose the smallest summary size such that increasing it does not result in changes in the best performing methods for either IL or IRL in the HIV simulator and PAC-MAN domains where this was possible. In the random gridworld domain, increasing the summary size always improved the performance of the IL methods, so we choose a summary size such that the best performing IRL methods did not change (HIV: 24; Gridworld: 24; PAC-MAN: 12).}
\label{fig:hyperparameters}
\end{figure}

We chose summary sizes and hyperparameter settings to analyze reconstruction quality by plotting the reconstruction quality of each hyperparameter setting for IL and IRL with the summary extracted using the corresponding model. Figure~\ref{fig:hyperparameters} shows this for 75 random restarts over summary sizes [12, 24, 36, 48, 60]; IL hyperparameters: kernel [RBF, polynomial], length scale [0.1, 1.] and degree [2, 3] (for polynomial kernel only); and IRL hyperparameters: trajectory lengths [1, 2, 3, 4]. The error bars correspond to 95\% confidence intervals.

For each reconstruction method, we additionally plotted the reconstruction quality with a random summary. For IL, we used only random summaries with trajectory length [1], and for the IRL, we tried random summaries with trajectory lengths [1, 2, 3, 4].

To determine the summary size $k$ for each domain, we chose the smallest summary size such that increasing it does not result in changes in the best performing methods for either IL or IRL in the HIV simulator and PAC-MAN domains where this was possible. In the random gridworld domain, increasing the summary size always improved the performance of the IL methods, so we choose a summary size such that the best performing IRL methods did not change (HIV: 24; Gridworld: 24; PAC-MAN: 12).  Our results in the main paper show only the best performing methods for IL and IRL at that summary size.

\section{User Study Summaries}
\label{sec:user_details}

We selected a single summary for each model in each domain, and a single set of 9 test questions for each domain in order to run the human-subject studies. We chose summaries and test states to have a similar pattern of high-accuracy reconstructions when the summarization model matches the extraction model, and low-accuracy reconstructions when the reconstruction model and the extraction model do not match. The summaries and test states we chose resulted in the reconstruction accuracies listed in Table~\ref{tab:gridworld_summaries}. 

The absolute values of the HIV summaries are different than those listed in Figure 1 of the main text because we used a shorter summary (12 states instead of 24) so that it could be easily visualized on a single page of the experiment. The accuracies computed on only the test states do not perfectly match the accuracies of the entire test grid, but they do preserve the pattern mentioned above of higher accuracies when reconstruction model and extraction model match, and lower otherwise. This allows us to test whether people's reconstructions have a difference in quality with different summaries.

\begin{table}[]
\centering
\begin{tabular}{ll|ll|ll}
\multicolumn{2}{c}{ } &            \multicolumn{2}{c}{HIV}  &  \multicolumn{2}{c}{Gridworld}  \\
            &               & IL   & IRL  &   IL   & IRL  \\ \hline 
\multirow{ 2}{*}{All States}  & IL            & 0.89 & 0.44 &  0.63 & 0.20 \\
            & IRL           & 0.76 & 0.91 & 0.24 & 0.91 \\ \hline 
\multirow{ 2}{*}{Test States} & IL            & 0.67 & 0.33 & 0.78 & 0.22 \\
            & IRL           & 0.33 & 0.78 & 0.11 & 1.  
\end{tabular}
\caption{Reconstruction accuracies of the summaries chosen for the human-subject studies on all points not included in the sumary (top rows) and on the set of 9 test states for which we asked people to predict actions (bottom rows). Rows represent the summary extraction model (IL/IRL), columns represent the reconstruction model (IL/IRL).}
\label{tab:gridworld_summaries}
\end{table}

\section{Human-Subject Study Interface}
\label{sec:user_interface}

\begin{figure}[H]
\centering
\includegraphics[width=1\columnwidth]{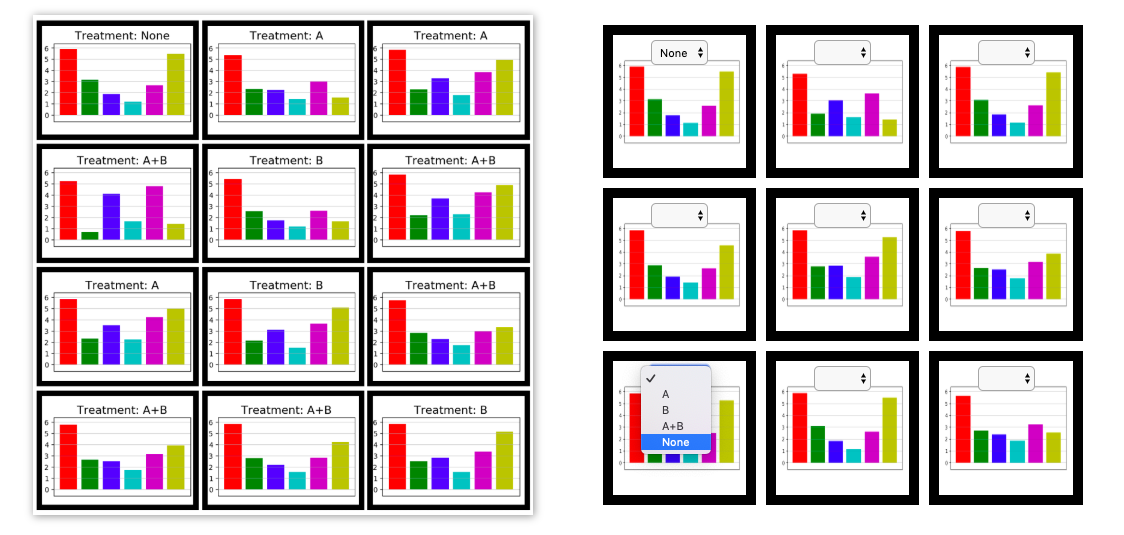}
\caption{The user study interface for an IL summary in the HIV simulator domain. The left side shows the summary and the right side shows the prediction states.}
\label{fig:interface_hiv_il}
\vspace{-0.2cm}
\end{figure}

\begin{figure}[H]
\centering
\includegraphics[width=1\columnwidth]{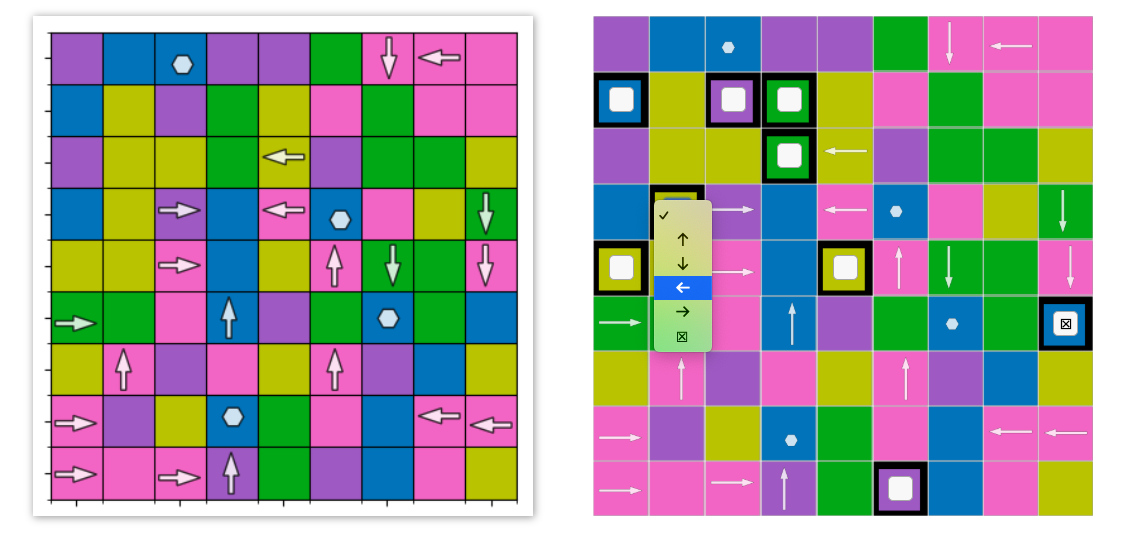}
\caption{The user study interface for an IL summary in the Gridworld domain. Both sides show the full grid, the left side shows the summary and the right side shows the the prediction states (tiles surrounded by black squares with a white box inside them) as well as the actions in the summary for convenience.}
\label{fig:interface_gridworld_il}
\vspace{-0.2cm}
\end{figure}

\begin{figure}[H]
\centering
\includegraphics[width=1\columnwidth]{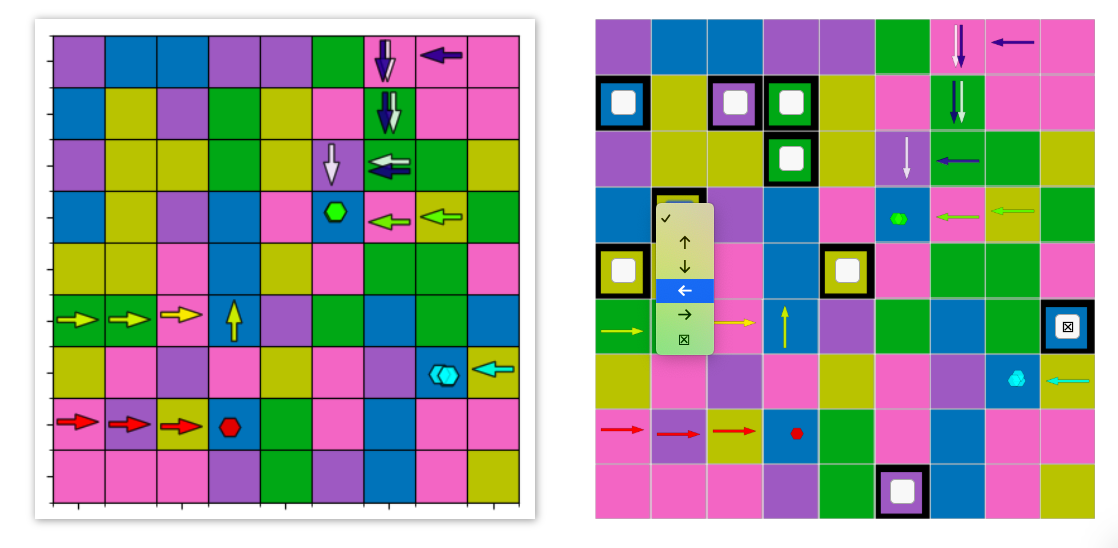}
\caption{The user study interface for an IRL summary in the Gridworld domain. Both sides show the full grid, the left side shows the summary and the right side shows the the prediction states (tiles surrounded by black squares with a white box inside them) as well as the actions in the summary for convenience. The summary displays 6 trajectories of 4 actions, corresponding to the different colored arrows.}
\label{fig:interface_gridworld_irl}
\vspace{-0.2cm}
\end{figure}

We conducted a human subject study in two domains - Gridworld and HIV simulator. Each participant viewed one summary extracted using IL or IRL, for one of the domains. During the task, participants were presented with a summary, and asked to predict the actions of the agent in states not shown in the summary. Figures \ref{fig:interface_hiv_il}, \ref{fig:interface_gridworld_il}, \ref{fig:interface_gridworld_irl} present screenshots of the different interfaces. In all interfaces, participants were asked to select the actions for the prediction states using a dropdown list. 

\section{Human-Subject Study Qualitative Responses}
\label{sec:user_qualitative}

After completing the prediction task, participants were asked to answer the following question: ``How did you decide which action the agent will take in each scenario?''. The purpose of this question was eliciting the reconstruction model people used while viewing the summary and making predictions. 

Based on the responses, we coded participants as using IL, IRL or not obviously using either. Among the IL responses we identified a few types of feature spaces to which participants referred. 

Table \ref{hiv_responses} and Table \ref{gridworld_responses} present the full set of responses for all participants in HIV simulator and Gridworld domains, respectively. These tables include the type of summary shown to the participant, the text of the response and the corresponding response type code. Table \ref{hiv_response_legend} and Table \ref{gridworld_response_legend} show the legend for the response type codes for HIV simulator and Gridworld domains, respectively. 

\begin{table}[h]
\centering
\begin{tabular}{ |c|c| }
\hline
 Code & Description \\ \hline
 IL-1 & IL - general similarity \\ 
 IL-2 & IL - absolute sizes of bars \\
 IL-3 & IL - relative bar levels \\
 IRL & IRL oriented \\
   -  & No obvious reconstruction \\
 \hline
\end{tabular}
\caption{Response types by reconstruction - HIV Simulator.
\label{hiv_response_legend}}
\end{table}

\begin{center}
\topcaption{HIV Simulator qualitative responses encoded by response type.\label{hiv_responses}}
\tablefirsthead{\hline\multicolumn{3}{| c |}{\textbf{HIV Simulator Qualitative Responses}}\\[0.5ex]\hline\textbf{Summary} & \textbf{Response Text} & \textbf{Response Type}\\[0.5ex]\hline}
\tablehead{\hline\multicolumn{3}{| c |}{\textbf{HIV Simulator Qualitative Responses}}\\[0.5ex]\hline\textbf{Summary} & \textbf{Response Text} & \textbf{Response Type}\\[0.5ex]}
\tabletail{\hline}
\begin{supertabular}{ | m{0.6in} | m{1.75in}| m{0.6in} | }
IL & I thought about the level that was displayed in the graph and then made my prediction based on the trend that I thought would follow. & IL-1 \\ \hline
IL & I tried to match the blood test to the examples. & IL-1 \\ \hline
IL & I just tried to find the graph that looked closest to the scenario and then used that to inform my decision & IL-1 \\ \hline
IL & I tried my best to find a pattern in the previous patient scenarios and use that to try to determine the treatment that the agent would choose. It was very difficult for me to find a pattern. Some of them, I just used gut instinct and went with what I thought was close to previous treatments chosen. & IL-1 \\ \hline
IL & I compared with the tests on left. & IL-1 \\ \hline
IRL & I matched it to the treatments on the left and picked the most simialr ones & IL-1 \\ \hline
IRL & I looked for a pattern in each treatment and tried to match it with the given scenario. & IL-1 \\ \hline
IL & I tried to match the blood tests that characterize the disease with the AI agent treatment. Each one had distinguishing characteristics. & IL-1 \\ \hline
IRL & Based off the patterns I saw in his previous treatments. & IL-1 \\ \hline
IRL & I mostly tried to match up the graph to the closest one in the summary, and chose the treatment it displayed. & IL-1 \\ \hline
IL & I compared the graphs of the scenarios against the treatment array I was given and chose the closest match. & IL-1 \\ \hline
IL & Tried to match to a similar patient on the left & IL-1 \\ \hline
IL & I examined the charts to see if there was one that corelated well, and then compared all of that letter to it to see if it was a common reaction. & IL-1 \\ \hline
IRL & Following the previous examples. & IL-1 \\ \hline
IL & Comparing the results to what was shown & IL-1 \\ \hline
IRL & I based it on the sequence that was presented to me. I looked at the colors of the graph and how much of each was shown and tried to match that with what I thought the agent would do next.  & IL-1 \\ \hline
IL & I tried to match up the graphs as best as I could. & IL-1 \\ \hline
IL & By assessing their blood tests and looking at prior treatments taken in similar blood tests before.  & IL-1 \\ \hline
IRL & I compared the sizes of the bars with the examples that were on the left and picked a treatment that had a similar result. & IL-1 \\ \hline
IL & I compared the images on the right with one on the left and chose the treatment based on similarity. & IL-1 \\ \hline
IRL & I tried to match them the best way I could by comparing the other scenarios.  & IL-1 \\ \hline
IL & Tried to find the closest scenario in the summary & IL-1 \\ \hline
IRL & I saw the graphs and how they looked with certain treatments and copied that & IL-1 \\ \hline
IL & I tried to find the samples that had very similar criteria for each of the blood test, including relative levels vs others as well as over all levels. & IL-1 \\ \hline
IL & Tried to find similarities in the graphs and make decisions accordingly. & IL-1 \\ \hline
IL & I chose based on the similarity of the blood tests levels from the scenarios on the left. & IL-1 \\ \hline
IL & I compared to the options given. & IL-1 \\ \hline
IRL & I compared the blood tests taken in each scenario from a previous one, and tried to replicate it. & IL-1 \\ \hline
IL & I tried to look for similarities between the completed graph and the graph that I had to fill in. If I noticed an inverse parabola, I would choose (top left). If I noticed that pink was above the tannish-yellow, I would try to correlate the degree to which it was higher and look at a similar completed graph. I tried to model my predictions of the agent's behavior based on the data that was already presented to me. & IL-1 \\ \hline
IL & I tried my best to compare the graphs & IL-1 \\ \hline
IRL & I comapred the graphs to the other graphs. & IL-1 \\ \hline
IL & I compared each set of graphs to the graphs for each treatment and tried to match similar graphs & IL-1 \\ \hline
IL & I tried to match the graphs with the treatments. For example if most of the A+B treatments seemed to form a curve on the graph I chose A+B if the graph has a curve. & IL-1 \\ \hline
IL & I tried to compare them to the similarities as shown on the left. & IL-1 \\ \hline
IRL & I looked at how similar a graph looked to one the agent treated and chose the same action that the agent chose. & IL-1 \\ \hline
IL & I tried to match the graphs with what the other had done/figure out key components of the chart. & IL-1 \\ \hline
IL & By using the examples and trying to match them accordingly. & IL-1 \\ \hline
IRL & I looked at each scenario and looked at the summary, I chose the one that I thought was similar to one that was in the summary. & IL-1 \\ \hline
IL & i looked at the same graphs & IL-1 \\ \hline
IL & I tried to match the levels with the graphs to those in the known actions, as best I could. & IL-1 \\ \hline
IRL & I compared the charts and chose the most similar treatment to the the patterns. & IL-1  \\ \hline
IRL & I was trying to look at the color graphs to compare the size & IL-2 \\ \hline
IL & Based on the values of each of the bars. & IL-2 \\ \hline
IL & I looked for patterns within the blood tests like the disparity between them. & IL-3 \\ \hline
IRL & I compared the charts and their results to the scenario I was evaluating. It seemed like there were some patterns in the bars based on the agents actions, and I tried to identify those patters as best I could. & IL-3 \\ \hline
IL & I chose the one that was closest to the levels.  & IL-3 \\ \hline
IL & I compared the size and order of the bar graphs, then guessed based on similarity. & IL-3 \\ \hline
IL & Comparative blood test tick levels & IL-3 \\ \hline
IRL & I tried to match the levels of the blood tests with another patient as much as possible and chose the same action as that one. I looked for key identifiers that separated the graph from others, such as two of the same level. & IL-3 \\ \hline
IRL & i compared the levels of each of the colors  and the shapes of the graphs.  & IL-3 \\ \hline
IRL & Relative similarity of bar heights in relationship to each-other,  inferred from previous actions.   & IL-3 \\ \hline
IRL & I tried to follow the patterns of the treatments along with the images provided and picked which was closer in relation.  & IL-4 \\ \hline
IL & Looking at past test and comparing which test were high or low and like these new tests. & IL-3 \\ \hline
IRL & I looked at the light blue/4th bar and compared how high it is to the examples. Then I looked at the dark blue bar/3rd bar, and compared also. I noticed that there were some differences in a+b or just A or B depending on how big those bars where. Plus the last bar usually indicates it's going to be a+b if its shorter. & IL-3 \\ \hline
IRL & Treatment A is used to decrease middle ones(blue, light blue and purple) and to increase light green which at the end. Treatment  is used to increase green which is second. & IRL \\ \hline
IRL & I based my answers solely on the results of each blood test. If the results were better for Treatment A, I would chose A.  & - \\ \hline
IRL & I tried to see the difference in the way the graphs moved & - \\ \hline
IL & Based on their policy & - \\ \hline
IRL & I compared it to the diagram. & - \\ \hline
IL & I just went with my intuition with what little understanding I had on how this works. & - \\ \hline
IL & this is very critical situation to find & - \\ \hline
IL & unsure what each bar represents (good or bad?), so just trying everything to see what works & - \\ \hline
IRL & The deontological class of ethical theories states that people should adhere to their obliga- tions and duties when engaged in decision making when ethics are in play. & - \\ \hline
IRL & I basically gave up & - \\ \hline
IRL & I tried to track the treatments with the bar chart & - \\ \hline
IL & if achieve in the mainly trying to reach this one. it is a treatment behavior on that ones. & - \\ \hline
IRL & Based on historical action highlighted in the bar graphs. And some luck was involved because more important data (actual numbers) was withheld.  & - \\ \hline
IL & I chose my actions based on the each patient's blood test in each scenario.  & - \\ \hline
IRL & I tried to decide based on the outcomes shown on the given charts. The most likely to be successful. & - \\ \hline
IRL & The graph results, shape and color & - \\ \hline
IRL & I tried to compare what was used. & - \\ \hline
IRL & what seemed best & - \\ \hline
\end{supertabular}
\end{center}

\begin{table}[h]
\centering
\begin{tabular}{ |c|c| }
\hline
 Code & Description \\ \hline
 IL-1 & IL - actions in similar tiles \\ 
 IL-2 & IL - direction \\
 IL-3 & IL - frequent action for color \\
 IL-4 & IL - surrounding tiles \\
 IRL  & IRL - goal oriented \\
   -  & No obvious reconstruction \\
 \hline
\end{tabular}
\caption{Response types by reconstruction - Gridworld.
\label{gridworld_response_legend}}
\end{table}
 
\begin{center}
\topcaption{Gridworld qualitative responses encoded by response type. \label{gridworld_responses}}
\tablefirsthead{\hline\multicolumn{3}{| c |}{\textbf{Gridworld Qualitative Responses}}\\[0.5ex]\hline\textbf{Summary} & \textbf{Response Text} & \textbf{Response Type}\\[0.5ex]\hline}
\tablehead{\hline\multicolumn{3}{| c |}{\textbf{Gridworld Qualitative Responses}}\\[0.5ex]\hline\textbf{Summary} & \textbf{Response Text} & \textbf{Response Type}\\[0.5ex]}
\tabletail{\hline}
\begin{supertabular}{ | m{0.6in} | m{1.75in}| m{0.6in} | }
IL & I thought they would act similarly to other tiles of the same color. & IL-1 \\ \hline
IL & I looked at what action was being taken in similar colored blocks and used that information to help me make my judgments. & IL-1 \\ \hline
IRL & based on actions took in tiles of the same color & IL-1 \\ \hline
IRL & I looked at what each color arrow did in each colored tile. & IL-1 \\ \hline
IRL & I tried to follow what I felt the directional pattern was. & IL-2 \\ \hline
IL & I tried to see what direction most of the arrows where taking and tried to see if it would be appropriate to follow the same position as the arrows already shown.  & IL-2 \\ \hline
IL & I looked at all the arrows and most of them in the same columns were going the same direction, so that's what I chose & IL-2 \\ \hline
IL & I tried comparing the colors and deciding which was more frequent for a color.  & IL-3 \\ \hline
IL & I tried to guess based on the colors and directions that seemed to be favored in the arrows displayed on the summary. & IL-3 \\ \hline
IRL & I was trying to be consistent with what colors had what action on it. & IL-3 \\ \hline
IL & I tried to use the nearest tile as a guide. & IL-4 \\ \hline
IL & I decided by the surrounding arrows If most were up I would click on the up arrow & IL-4 \\ \hline
IL & By looking at the summery and seeing which way the marked tiles were moving around the marked tile i thought it was more likely to follow the same pattern. & IL-4 \\ \hline
IL & I looked at the surrounding block and went with the most common move. If there was a surrounding block that was also the same color then that took priority. & IL-4 \\ \hline
IL & I tried looking to see where the majority of the arrows were going to, and chose those to move to, which seemed to be blue and green. & IRL \\ \hline
IL & I figured each section would have a beginning and lead to a stop square, so I marked the square accordingly.  & IRL \\ \hline
IL & I was trying to make the fastest path to a blue square & IRL \\ \hline
IL & It seems like the agent tries to get to blue in as few moves as possible, and then stops there. I chose based on that reasoning. & IRL \\ \hline
IL & I chose the one that I thought would get them to the higher point squares or blue. & IRL \\ \hline
IL & I tried to get it to go to a blue spot, because I think that's what it wanted & IRL \\ \hline
IL & The first time I was just attempting to do something similar to the given board, however, the second time I ranked the colors.  & IRL \\ \hline
IRL & Which would take them to where there was a stop & IRL \\ \hline
IRL & I tried to predict how they could end up at the stops in place. & IRL \\ \hline
IRL & The agent seems to prefer to stop on blue squares, so I chose actions that led to blue squares. & IRL \\ \hline
IRL & I generally assumed it would take the shortest path to a blue square and guessed accordingly. & IRL \\ \hline
IRL & I pushed them in the shortest direction to a blue square, or stopped if they were already on one. & IRL \\ \hline
IRL & I decided that the computer seems to be always working towards a blue square. I chose the simplest path to get to a blue square.  & IRL \\ \hline
IL & The agent seems primarily motivated to getting to the blue squares and stopping. So I selected directions that would move the agent toward the blue squares. & IRL \\ \hline
IRL & The agent chooses the shortest path to the nearest blue square. & IRL \\ \hline
IRL & I figured that the agent was attempting to get to (and stay on) the blue squares, so I assumed the agent would take the fewest steps possible to reach a blue square. I assumed it would take the action that would get it closest to a blue square. & IRL \\ \hline
IRL & I think blue is always stop. It seems the agent travels to a stop. There may be a task to perform at that spot. & IRL \\ \hline
IRL & I tried to make a path according to the other arrows. I figured they run a few squares before they stop. Seemed they stopped in blue square so made one a stop in the blue square. & IRL \\ \hline
IRL & I decided that the agent would take the shortest route available to the nearest blue tile.  & IRL \\ \hline
IRL & I thought the agent was trying to go to the nearest blue square and stop there. It seemed to prefer to go sideways but I could not decide if it preferred to go left or right or toward the center of the diagram. & IRL \\ \hline
IL & I decided which action the agent would take based on the amount of room that they had, which way the other agents were going, and how they could get the most points. & - \\ \hline
IL & Honestly, just guessing. & - \\ \hline
IL & I honestly have no idea, except that I followed the behavior of the closest tile with a behavior either in a column or in a row. & - \\ \hline
IL & I looked at the arrows already there as well as the stops and tried to ascertain the clearer path. & - \\ \hline
IL & To be honest I guessed in an intuitive manner, I can not really explain it. & - \\ \hline
IL & If there was a color tile that matched in a way that it could move. & - \\ \hline
IL & I just looked at where the arrows were headed and guessed the direction he wanted to go. I made sure not to overlap existing moves. & - \\ \hline
IL & I completely guessed. & - \\ \hline
IL & Just kind of guessed I guess. Didn't want them to cross any paths or back track. & - \\ \hline
IL & I looked at the surrounding movements and saw what would make sense for it to get there. & - \\ \hline
IL & I went with what appeared to be the most logical steps and also took into account what the board was displaying & - \\ \hline
IL & I guess it was just a gut feeling and trying to find patterns. & - \\ \hline
IRL & Some arrows and directions matched up to a sequence. Others, were a guess. & - \\ \hline
IRL & I observed the agents steps taken in the last scenario. & - \\ \hline
IRL & I tried to see where the arrows were going to and to follow a similar pattern. & - \\ \hline
IRL & I tried to visualize how the agent maybe would naturally be moving. & - \\ \hline
IRL & i predicted the action s that will be taken by the agent in the maptiles & - \\ \hline
IRL & tried to see the sequence & - \\ \hline
IRL & I looked at the directions he already traveled in the color I was judging and tried to think about where it might go next. & - \\ \hline
IL & I tried to intuit which decision would be the most effective & - \\ \hline
IL & I tried to imagine what the most logical move would be for the agent. & - \\ \hline
IL & I looked at what they did in the squares around it & - \\ \hline
IL & I looked at the arrows around the boxes & - \\ \hline
IL & I looked at the surrounding colors and details of the arrows/stops and chose around that & - \\ \hline
IL & I looked at the color of the squares they mostly preferred to move toward, then decided how they would act based on that. & - \\ \hline
IRL & I made a guess based on the actions in the surrounding squares. & - \\ \hline
IRL & guessed & - \\ \hline
IRL & I went with what the agent had done the precious 3 turns & - \\ \hline
IRL & I tried to use the image on the left for guidance but ultimately went with what made since, based on the position of the white square. & - \\ \hline
IRL & I just made my choices by what would make sense on the grid & - \\ \hline
IRL & I followed the sequebce of the arrow. & - \\ \hline
IRL & It seems a reasonable way to go based on the potential movements.  & - \\ \hline
IRL & I just tried to think what would be most logical. & - \\ \hline
IRL & It seem like the best case scenario with a little guess work thrown in. & - \\ \hline
IRL & i thought about the way that made the most sense and choice that & - \\ \hline
IRL & I looked at the other summaries and tried to find a pattern and based it on that & - \\ \hline
IRL & look at what it did before when faced with options it faces now based on color available in the legal moves & - \\ \hline
IRL & I looked at the tiles to see what he mostly was navigating towards, in favor over another choice. & - \\ \hline
IRL & I tried to predict the next movement as closely as I could. If it seemed likely, then I decided on that specific action. & - \\ \hline
IRL & I think I just went with following the general consensus of where the other agents nearby seemed to be headed. & - \\ \hline
IRL & Just a gut feeling honestly. & - \\ \hline
\end{supertabular}
\end{center}

\end{document}